\documentclass[times,twocolumn,final,authoryear]{elsarticle}

\usepackage{prletters}
\usepackage{framed,multirow}

\usepackage{amssymb}
\usepackage{latexsym}

\usepackage{url}
\usepackage{xcolor}
\definecolor{newcolor}{rgb}{.8,.349,.1}
\usepackage{hyperref}
\usepackage{multirow}
\usepackage{multicol}
\usepackage{graphicx}
\usepackage{subfigure}
\usepackage{bbding}
\usepackage{amsmath}
\usepackage{amssymb}
\usepackage{epstopdf}
\usepackage{siunitx}
\usepackage{amsthm}

\usepackage{booktabs}
\usepackage{algorithm}
\usepackage{algorithmic}
\usepackage{diagbox}
\usepackage{chngpage}
\usepackage{tablefootnote}
\usepackage{tabularx}
\usepackage{adjustbox}
\usepackage{enumitem}
\journal{Pattern Recognition Letters}

\begin{document}

\begin{table*}[!t]
	\ifpreprint\else\vspace*{-15pc}\fi
	
	\section*{Research Highlights (Required)}
	
	To create your highlights, please type the highlights against each
	\verb+\item+ command. 
	
	\vskip1pc
	
	\fboxsep=6pt
	\fbox{
		\begin{minipage}{.95\textwidth}
			It should be short collection of bullet points that convey the core
			findings of the article. It should include 3 to 5 bullet points
			(maximum 85 characters, including spaces, per bullet point.)  
			\vskip1pc
			\begin{itemize}
				
				\item We survey deep learning based HAR in sensor modality, deep model, and application.
				
				\item We comprehensively discuss the insights of deep learning models for HAR tasks.
				
				\item We extensively investigate why deep learning can improve the performance of HAR.
				
				\item We also summarize the public HAR datasets frequently used for research purpose.
				
				\item We present some grand challenges and feasible solutions for deep learning based HAR.
				
			\end{itemize}
			\vskip1pc
		\end{minipage}
	}
	
\end{table*}

\clearpage

\ifpreprint
\setcounter{page}{1}
\else
\setcounter{page}{1}
\fi

\begin{frontmatter}
	
\title{Deep Learning for Sensor-based Activity Recognition: A Survey}

\author[1,2]{Jindong \snm{Wang}} 

\author[1,2]{Yiqiang \snm{Chen}\corref{cor1}}
\ead{yqchen@ict.ac.cn}
\cortext[cor1]{Corresponding author}
\author[3]{Shuji \snm{Hao}}
\author[1,2]{Xiaohui \snm{Peng}}
\author[1,2]{Lisha \snm{Hu}}

\address[1]{Beijing Key Laboratory of Mobile Computing and Pervasive Device, Institute of Computing Technology, Chinese Academy of Sciences, Beijing, China}
%\address[2]{, Beijing, China}
\address[2]{University of Chinese Academy of Sciences, Beijing, China}
\address[3]{Institute of High Performance Computing, A*STAR, Singapore}

\received{1 May 2013}
\finalform{10 May 2013}
\accepted{13 May 2013}
\availableonline{15 May 2013}
\communicated{S. Sarkar}

\begin{abstract}
	Sensor-based activity recognition seeks the profound high-level knowledge about human activities from multitudes of low-level sensor readings. Conventional pattern recognition approaches have made tremendous progress in the past years. However, those methods often heavily rely on heuristic hand-crafted feature extraction, which could hinder their generalization performance. Additionally, existing methods are undermined for unsupervised and incremental learning tasks. Recently, the recent advancement of deep learning makes it possible to perform automatic high-level feature extraction thus achieves promising performance in many areas. Since then, deep learning based methods have been widely adopted for the sensor-based activity recognition tasks. This paper surveys the recent advance of deep learning based sensor-based activity recognition. We summarize existing literature from three aspects: sensor modality, deep model, and application. We also present detailed insights on existing work and propose grand challenges for future research.
	\\
	\textit{Keywords:} Deep learning; activity recognition; pattern recognition; pervasive computing
\end{abstract}

%\begin{keyword}
%	\MSC 41A05\sep 41A10\sep 65D05\sep 65D17
%	\KWD Deep learning\sep Activity recognition\sep Ubiquitous computing
%	
%	%% MSC codes here, in the form: \MSC code \sep code
%	%% or \MSC[2008] code \sepc code (2000 is the default)
%\end{keyword}
	
\end{frontmatter}

%\linenumbers

%% main text
\section{Introduction}
\label{sec-intro}
Human activity recognition~(HAR) plays an important role in people's daily life for its competence in learning profound high-level knowledge about human activity from raw sensor inputs. Successful HAR applications include home behavior analysis~\citep{vepakomma2015wristocracy}, video surveillance~\citep{qin2016compressive}, gait analysis~\citep{hammerla2016deep}, and gesture recognition~\citep{kim2016hand}. There are mainly two types of HAR: \textit{video-based} HAR and \textit{sensor-based} HAR~\citep{cook2013transfer}. Video-based HAR analyzes videos or images containing human motions from the camera, while sensor-based HAR focuses on the motion data from smart sensors such as an accelerometer, gyroscope, Bluetooth, sound sensors and so on. Due to the thriving development of sensor technology and pervasive computing, sensor-based HAR is becoming more popular and widely used with privacy well protected. Therefore, in this paper, our main focus is on sensor-based HAR.

%As a classic pattern recognition (PR) problem, HAR consists of critical procedures like feature extraction and classification model building. With representative knowledge extracted from raw inputs, a classification model can be built using those knowledge to perform activity recognition.

HAR can be treated as a typical pattern recognition~(PR) problem. Conventional PR approaches have made tremendous progress on HAR by adopting machine learning algorithms such as decision tree, support vector machine, naive Bayes, and hidden Markov models~\citep{lara2013survey}. It is no wonder that in some controlled environments where there are only a few labeled data or certain domain knowledge is required~(e.g. some disease issues), conventional PR methods are fully capable of achieving satisfying results. However, in most daily HAR tasks, those methods may heavily rely on heuristic hand-crafted feature extraction, which is usually limited by human domain knowledge~\citep{bengio2013deep}. Furthermore, only shallow features can be learned by those approaches~\citep{yang2015deep}, leading to undermined performance for unsupervised and incremental tasks. Due to those limitations, the performances of conventional PR methods are restricted regarding classification accuracy and model generalization.

Recent years have witnessed the fast development and advancement of deep learning, which achieves unparalleled performance in many areas such as visual object recognition, natural language processing, and logic reasoning~\citep{lecun2015deep}. Different from traditional PR methods, deep learning can largely relieve the effort on designing features and can learn much more high-level and meaningful features by training an end-to-end neural network. In addition, the deep network structure is more feasible to perform unsupervised and incremental learning. Therefore, deep learning is an ideal approach for HAR and has been widely explored in existing work~\citep{lane2015deepear,alsheikh2015deep,plotz2011feature}. 

Although some surveys have been conducted in deep learning~\citep{lecun2015deep,schmidhuber2015deep,bengio2013deep} and HAR~\citep{lara2013survey,bulling2014tutorial}, respectively, there has been no specific survey focusing on the intersections of these two areas. To our best knowledge, this is the \textit{first} article to present the recent advance on deep learning based HAR. We hope this survey can provide a helpful summary of existing work, and present potential future research directions.

%To be short, the contributions of this paper are as follows:
%\begin{itemize}
%	\item We review and summarize the recent advance of deep learning based HAR from three aspects: sensor modality, deep learning models, and applications.
%	\item We comprehensively discuss the recent advancement of deep learning model for HAR tasks.
%	\item We extensively present and discuss some grand challenges for deep learning based HAR for future research direction.
%\end{itemize}

The rest of this paper is organized as follows. In Section~\ref{sec-background}, we briefly introduce sensor-based activity recognition and explain why deep learning can improve its performance. In Section~\ref{sec-modalilty}, \ref{sec-model} and~\ref{sec-app}, we review recent advance of deep learning based HAR from three aspects: sensor modality, deep model, and application, respectively. We also introduce several benchmark datasets. Section~\ref{sec-summary} presents summary and insights on existing work. In Section~\ref{sec-discuss}, we discuss some grand challenges and feasible solutions. Finally, this paper is concluded in Section~\ref{sec-conclusion}.

\begin{figure}[tb]
	\centering
	\setlength{\fboxrule}{1pt}   
	\setlength{\fboxsep}{0.1cm}
	\fbox{\includegraphics[scale=0.38]{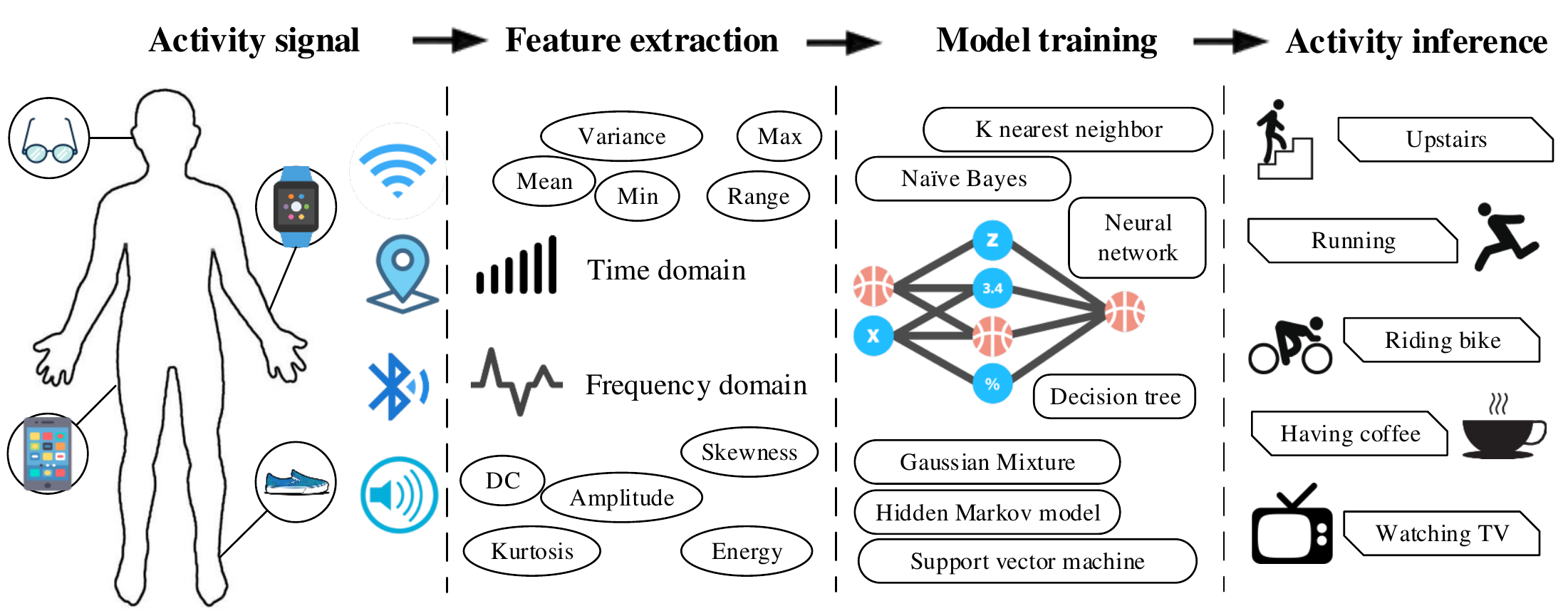}}
	\caption{An illustration of sensor-based activity recognition using conventional pattern recognition approaches.}
	\label{fig-ar}
	\vspace{-.1in}
\end{figure}

\section{Background}
\label{sec-background}

\subsection{Sensor-based Activity Recognition}
HAR aims to understand human behaviors which enable the computing systems to proactively assist users based on their requirement~\citep{bulling2014tutorial}. Formally speaking, suppose a user is performing some kinds of activities belonging to a predefined activity set $A$:
\begin{equation}
A=\{A_i\}^{m}_{i=1}
\end{equation}
where $m$ denotes the number of activity types. There is a sequence of sensor reading that captures the activity information
\begin{equation}
\mathbf{s}=\{\mathbf{d}_1,\mathbf{d}_2,\cdots,\mathbf{d}_t,\cdots \mathbf{d}_n\}
\end{equation}
where $\mathbf{d}_t$ denotes the sensor reading at time $t$. 

We need to build a model $\mathcal{F}$ to predict the activity sequence based on sensor reading $\textbf{s}$
\begin{equation}
\hat{A}=\{\hat{A}_j\}^{n}_{j=1}=\mathcal{F}(\mathbf{s}), \quad \hat{A}_j \in A
\end{equation}
while the true activity sequence (ground truth) is denoted as
\begin{equation}
A^\ast = \{A^\ast_j\}^{n}_{j=1}, \quad A^\ast_j \in A
\end{equation}
where $n$ denotes the length of sequence and $n \ge m$.

The goal of HAR is to learn the model $\mathcal{F}$ by minimizing the discrepancy between predicted activity $\hat{A}$ and the ground truth activity $A^\ast$. Typically, a positive loss function $\mathcal{L}(\mathcal{F}(\textbf{s}),A^\ast)$ is constructed to reflect their discrepancy. $\mathcal{F}$ usually does not directly take $\textbf{s}$ as input, and it usually assumes that there is a projection function $\Phi$ that projects the sensor reading data $\mathbf{d_i} \in \mathbf{s}$ to a $d$-dimensional feature vector $\Phi(\textbf{d}_i) \in \mathbb{R}^d$. To that end, the goal turns into minimizing the loss function $\mathcal{L}(\mathcal{F}(\Phi(\textbf{d}_i)),A^\ast)$.

%So the key to success of HAR is to obtain advanced transformation function $\Phi(\cdot)$ and the model function $\mathcal{F}$. 

Fig.~\ref{fig-ar} presents a typical flowchart of HAR using conventional PR approaches. First, raw signal inputs are obtained from several types of sensors~(smartphones, watches, Wi-Fi, Bluetooth, sound etc.). Second, features are manually extracted from those readings based on human knowledge~\citep{bao2004activity}, such as the \textit{mean}, \textit{variance}, \textit{DC}, and \textit{amplitude} in traditional machine learning approaches~\citep{hu2016less}. Finally, those features serve as inputs to train a PR model to make activity inference in real HAR tasks. 

\subsection{Why Deep Learning?}

Conventional PR approaches have made tremendous progress in HAR~\citep{bulling2014tutorial}. However, there are several drawbacks to conventional PR methods. 

Firstly, the features are always extracted via a heuristic and hand-crafted way, which heavily relies on human experience or domain knowledge. This human knowledge may help in certain task-specific settings, but for more general environments and tasks, this will result in a lower chance and longer time to build a successful activity recognition system. 

Secondly, only shallow features can be learned according to human expertise~\citep{yang2015deep}. Those shallow features often refer to some statistical information including mean, variance, frequency and amplitude etc. They can only be used to recognize low-level activities like \textit{walking} or \textit{running}, and hard to infer high-level or context-aware activities~\citep{yang2009activity}. For instance, \textit{having coffee} is more complex and nearly impossible to be recognized by using only shallow features.

Thirdly, conventional PR approaches often require a large amount of well-labeled data to train the model. However, most of the activity data are remaining unlabeled in real applications. Thus, these models' performance is undermined in unsupervised learning tasks~\citep{bengio2013deep}. In contrast, existing deep generative networks~\citep{hinton2006fast} are able to exploit the unlabeled samples for model training. 

Moreover, most existing PR models mainly focus on learning from static data; while activity data in real life are coming in stream, requiring robust online and incremental learning. 

\begin{figure}[tb]
	\centering
	\setlength{\fboxrule}{1pt}   
	\setlength{\fboxsep}{0.1cm}
	\fbox{\includegraphics[scale=0.38]{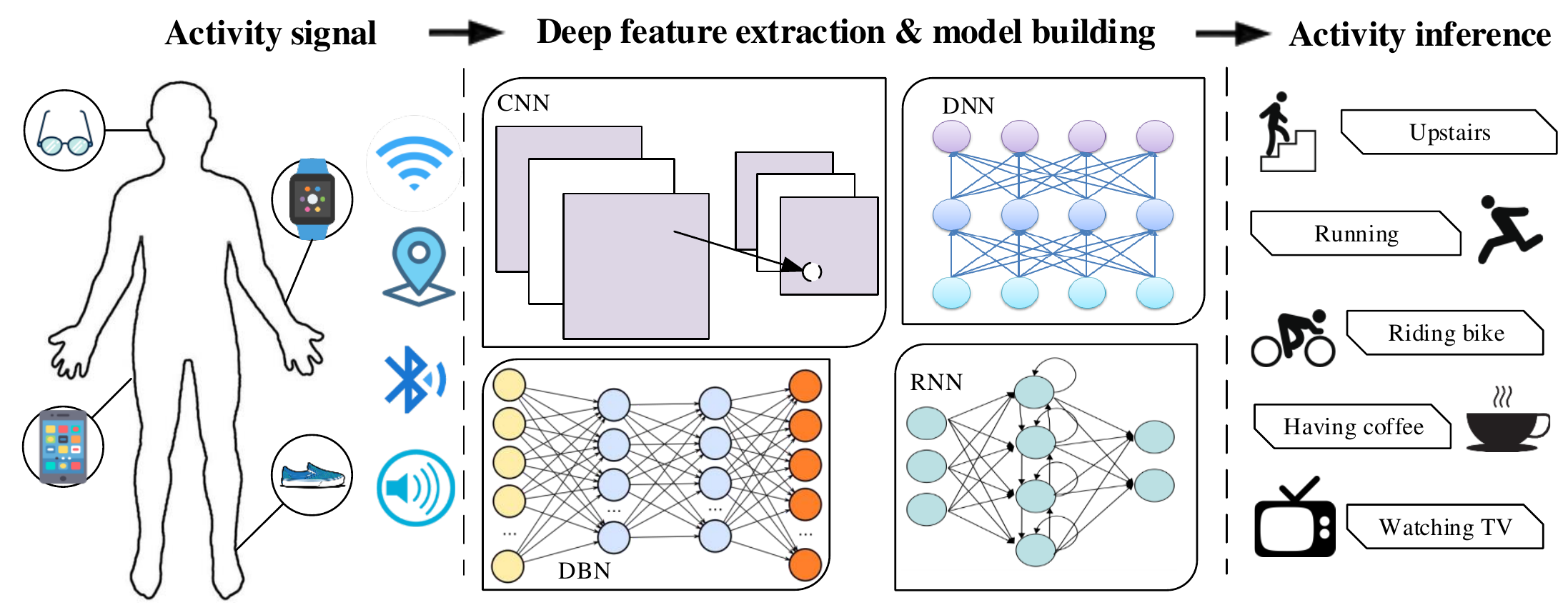}}
	\caption{An illustration of sensor-based activity recognition using deep learning approaches.}
	\label{fig-deepar}
	\vspace{-.1in}
\end{figure}

Deep learning tends to overcome those limitations. Fig.~\ref{fig-deepar} shows how deep learning works for HAR with different types of networks. Compared to Fig.~\ref{fig-ar}, the feature extraction and model building procedures are often performed simultaneously in the deep learning models. The features can be learned automatically through the network instead of being manually designed. Besides, the deep neural network can also extract high-level representation in deep layer, which makes it more suitable for complex activity recognition tasks. When faced with a large amount of unlabeled data, deep generative models~\citep{hinton2006fast} are able to exploit the unlabeled data for model training. What's more, deep learning models trained on a large-scale labeled dataset can usually be transferred to new tasks where there are few or none labels.

%Moreover, the multi-layer structure also makes deep learning more competent to tackle online / incremental learning~\citep{bengio2013deep}.

%There are already several work using deep learning to improve the performance of HAR~\citep{alsheikh2015deep,lane2015deepear,chen2015deep}. 

In the following sections, we mainly summarize the existing work based on the pipeline of HAR: (a)~\textit{sensor modality}, (b)~\textit{deep model}, and (c)~\textit{application}. 

%These aspects are based on the pipeline of HAR: initially, sensors are used to capture activity information, then a deep model is trained, finally the well-trained model can be applied to real applications.

\section{Sensor Modality}
\label{sec-modalilty}
Although some HAR approaches can be generalized to all sensor modalities, most of them are only specific to certain types. According to~\citep{chavarriaga2013opportunity}, we mainly classify those modalities into three aspects: \textit{body-worn sensors}, \textit{object sensors}, and \textit{ambient sensors}. Table~\ref{tb-modality} briefly outlines all the modalities.
%For each modality, we first provide a brief introduction, then we review the related work. 

\subsection{Body-worn Sensor}
Body-worn sensors are one of the most common modalities in HAR. Those sensors are often worn by the users, such as an accelerometer, magnetometer, and gyroscope. The acceleration and angular velocity are changed according to human body movements; thus they can infer human activities. Those sensors can often be found on smart phones, watches, bands, glasses, and helmets. 

Body-worn sensors were widely used in deep learning based HAR~\citep{chen2015deep,plotz2011feature,zeng2014convolutional,jiang2015human,yang2015deep}. Among those work, the accelerometer is mostly adopted. Gyroscope and magnetometer are also frequently used together with the accelerometer. Those sensors are often exploited to recognize activities of daily living~(ADL) and sports. Instead of extracting statistical and frequency features from the movement data, the original signal is directly used as inputs for the network.

\begin{table*}[t!]
	\centering
	\caption{Sensor modalities for HAR tasks}
	\label{tb-modality}
	\resizebox{1\textwidth}{!}{
	\begin{tabular}{|l|l|l|}
		\hline
		\textbf{Modality} & \textbf{Description} & \textbf{Examples} \\ \hline \hline
		\textbf{Body-worn}& Worn by the user to describe the body movements & Smartphone, watch, or band's accelerometer, gyroscope etc.\\ \hline
		\textbf{Object} & Attached to objects to capture objects movements & RFID, accelerometer on cup etc. \\ \hline
		\textbf{Ambient} & Applied in environment to reflect user interaction & Sound, door sensor, WiFi, Bluetooth etc. \\ \hline
		\textbf{Hybrid} & Crossing sensor boundary & Combination of types, often deployed in smart environments \\ \hline
	\end{tabular}
}
\vspace{-.1in}
\end{table*}

\subsection{Object Sensor}
Object sensors are usually placed on objects to detect the movement of a specific object~\citep{chavarriaga2013opportunity}. Different from body-worn sensors which capture human movements, object sensors are mainly used to detect the movement of certain objects in order to infer human activities. For instance, the accelerometer attached to a cup can be used to detect the \textit{drinking water} activity. Radio frequency identifier~(RFID) tags are typically used as object sensors and deployed in smart home environment~\citep{vepakomma2015wristocracy,yang2015deep,fang2014recognizing} and medical activities~\citep{li2016deep2,wang2016human}. The RFID can provide more fine-grained information for more complex activity recognition.

It should be noted that object sensors are less used than body-worn sensors due to the difficulty in its deployment. Besides, the combination of object sensors with other types is emerging in order to recognize more high-level activities~\citep{yang2009activity}.

\subsection{Ambient Sensor}
Ambient sensors are used to capture the interaction between humans and the environment. They are usually embedded in users' smart environment. There are many kinds of ambient sensors such as radar, sound sensors, pressure sensors, and temperature sensors. Different from object sensors which measure the object movements, ambient sensors are used to capture the change of the environment.

Several literature used ambient sensors to recognize daily activities and hand gesture~\citep{lane2015deepear,wang2016human,kim2016hand}. Most of the work was tested in the smart home environment. Same as object sensors, the deployment of ambient sensors is also difficult. In addition, ambient sensors are easily affected by the environment, and only certain types of activities can be robustly inferred.

% With the proliferation of ubiquitous computing, there is a promising future that ambient sensors will be more widely used for human activity recognition.

\subsection{Hybrid Sensor}
\label{sec-cross}
Some work combined different types of sensors for HAR. As shown in \citep{hayashi2015daily}, combining acceleration with acoustic information could improve the accuracy of HAR. Ambient sensors are also used together with object sensors; hence they can record both the object movements and environment state. \citep{vepakomma2015wristocracy} designed a smart home environment called \textit{A-Wristocracy}, where a large number of fine-grained and complex activities of multiple occupants can be recognized through body-worn, object, and ambient sensors. It is obvious that the combination of sensors is capable of capturing rich information of human activities, which is also possible for a real smart home system in the future.

\section{Deep Model}
\label{sec-model}
In this section, we investigate the deep learning models used in HAR tasks. Table~\ref{tb-model} lists all the models.

%Inspired by~\citep{deng2014tutorial}, we categorize the related work into three categories: generative deep architecture, discriminative deep architecture and hybrid deep architecture. Table~\ref{tb-model} presents a brief description of all those three deep learning models for HAR tasks.

%\subsection{Discriminative Deep Architecture}
%Discriminative deep architecture offers discriminative power to patterns via characterizing the posterior distributions of classes conditioned on the visible data~\citep{deng2014tutorial}. Existing research mostly falls into two deep learning models: (a)~deep fully-connected network and (b)~convolutional neural network. 

\subsection{Deep Neural Network}
Deep neural network~(DNN) is developed from artificial neural network~(ANN). Traditional ANN often contains very few hidden layers~(shallow) while DNN contains more~(deep). With more layers, DNN is more capable of learning from large data. DNN usually serves as the dense layer of other deep models. For example, in a convolution neural network, several dense layers are often added after the convolution layers. In this part, we mainly focus on DNN as a single model, while in other sections we will discuss the dense layer.

\citep{vepakomma2015wristocracy} first extracted hand-engineered features from the sensors, then those features are fed into a DNN model. Similarly, \citep{walse2016pca} performed PCA before using DNN. In those work, DNN only served as a classification model after hand-crafted feature extraction, hence they may not generalize well. And the network was rather shallow. \citep{hammerla2016deep} used a 5-hidden-layer DNN to perform automatic feature learning and classification with improved performance. Those work indicated that, when the HAR data is multi-dimensional and activities are more complex, more hidden layers can help the model train well since their representation capability is stronger~\citep{bengio2013deep}. However, more details should be considered in certain situations to help the model fine-tune better.

%As large-scale data is collected, researchers are more keen to apply DNN to other deep models.

\subsection{Convolutional Neural Network}
Convolutional Neural Network~(ConvNets, or CNN) leverages three important ideas: sparse interactions, parameter sharing, and equivariant representations~\citep{lecun2015deep}. After convolution, there are usually pooling and fully-connected layers, which perform classification or regression tasks. 

CNN is competent to extract features from signals and it has achieved promising results in image classification, speech recognition, and text analysis. When applied to time series classification like HAR, CNN has two advantages over other models: local dependency and scale invariance. Local dependency means the nearby signals in HAR are likely to be correlated, while scale invariance refers to the scale-invariant for different paces or frequencies. Due to the effectiveness of CNN, most of the surveyed work focused on this area.

When applying CNN to HAR, there are several aspects to be considered: \textit{input adaptation}, \textit{pooling}, and \textit{weight-sharing}.

\textit{1) Input adaptation}. Unlike images, most HAR sensors produce \textit{time series} readings such as acceleration signal, which is temporal multi-dimensional 1D readings. Input adaptation is necessary before applying CNN to those inputs. The main idea is to adapt the inputs in order to form a \textit{virtual image}. There are mainly two types of adaptation: \textit{model-driven} and \textit{data-driven}. 

\begin{itemize}[noitemsep,nolistsep]
	\item \textit{Data-driven} approach treats each dimension as a channel, then performs 1D convolution on them. After convolution and pooling, the outputs of each channel are flattened to unified DNN layers. A very early work is~\citep{zeng2014convolutional}, where each dimension of the accelerometer was treated as one channel like RGB of an image, then the convolution and pooling were performed separately. \citep{yang2015deep} further proposed to unify and share weights in multi-sensor CNN by using 1D convolution in the same temporal window. Along with this line, \citep{chen2015deep} resized the convolution kernel to obtain the best kernel for HAR data. Other similar work include \citep{hammerla2016deep,sathyanarayana2016impact,pourbabaee2017deep}. This data-driven approach treats the 1D sensor reading as a 1D image, which is simple and easy to implement. The disadvantage of this approach is the ignorance of dependencies between dimension and sensors, which may influence the performance.
	
	\item \textit{Model-driven} approach resizes the inputs to a virtual 2D \textit{image} so as to adopt a 2D convolution. This approach usually pertains to non-trivial input tuning techniques. \citep{ha2015multi} combined all dimensions to form an image, while \citep{jiang2015human} designed a more complex algorithm to transform the time series into an image. In~\citep{singh2017transforming}, pressure sensor data was transformed to the image via modality transformation. Other similar work include~\citep{ravi2016deep,li2016deep2}. This model-driven approach can make use of the temporal correlation of sensor. But the map of time series to image is non-trivial task and needs domain knowledge.
	
\end{itemize}

\textit{2) Pooling}. The \textit{convolution-pooling} combination is common in CNN, and most approaches performed max or average pooling after convolution~\citep{ha2015multi,kim2016hand,pourbabaee2017deep}. Apart from avoiding overfitting, pooling can also speed up the training process on large data~\citep{bengio2013deep}.

\textit{3) Weight-sharing}. Weight sharing~\citep{zebin2016human,sathyanarayana2016impact} is an efficient method to speed up the training process on a new task. \citep{zeng2014convolutional} utilized a relaxed partial weight sharing technique since the signal appeared in different units may behave differently. \citep{ha2016convolutional} adopted a CNN-pf and CNN-pff structure to investigate the performance of different weight-sharing techniques. It is shown in those literature that partial weight-sharing could improve the performance of CNN.

%\subsection{Generative Deep Architecture}
%Generative deep architecture aims to build the model by characterizing joint distributions from the visible data and their classes. Although there are many types of deep models under this architecture, we mainly introduce three popular ones adopted in HAR tasks: (a)~autoencoder, (b)~restricted Boltzmann machine and (c)~recurrent neural network.

\begin{table*}[t!]
	\centering
	\caption{Deep learning models for HAR tasks}
	\label{tb-model}
	\begin{tabular}{|l|l|}
		\hline
		\textbf{Model} & \textbf{Description} \\ \hline \hline
		\textbf{DNN} & Deep fully-connected network, artificial neural network with deep layers \\ \hline
		\textbf{CNN} & Convolutional neural network, multiple convolution operations for feature extraction \\ \hline
		\textbf{RNN} & Recurrent neural network, network with time correlations and LSTM \\ \hline
		\textbf{DBN / RBM} & Deep belief network and restricted Boltzmann machine \\ \hline
		\textbf{SAE} & Stacked autoencoder, feature learning by decoding-encoding autoencoder \\ \hline
		\textbf{Hybrid} & combination of some deep models \\ \hline
	\end{tabular}
	\vspace{-.1in}
\end{table*}

\subsection{Autoencoder}
Autoencoder learns a latent representation of the input values through the hidden layers, which can be considered as an encoding-decoding procedure. The purpose of autoencoder is to learn more advanced feature representation via an unsupervised learning schema. Stacked autoencoder~(SAE) is the stack of some autoencoders. SAE treats every layer as the basic model of autoencoder. After several rounds of training, the learned features are stacked with labels to form a classifier.

\citep{almaslukh2017effective,wang2016human} used SAE for HAR, where they first adopted the greedy layer-wise pre-training~\citep{hinton2006fast}, then performed fine-tuning. Compared to those works, \citep{li2014unsupervised} investigated the sparse autoencoder by adding KL divergence and noise to the cost function, which indicates that adding sparse constraints could improve the performance of HAR. The advantage of SAE is that it can perform unsupervised feature learning for HAR, which could be a powerful tool for feature extraction. But SAE depends too much on its layers and activation functions which may be hard to search the optimal solutions.

\subsection{Restricted Boltzmann Machine}
Restricted Boltzmann machine (RBM) is a bipartite, fully-connected, undirected graph consisting of a visible layer and a hidden layer \citep{hinton2006fast}. The stacked RBM is called deep belief network~(DBN) by treating every two consecutive layers as an RBM. DBN/RBM is often followed by fully-connected layers.

In pre-training, most work applied Gaussian RBM in the first layer while binary RBM for the rest layers~\citep{plotz2011feature,hammerla2015pd,lane2015deepear}. For multi-modal sensors, \citep{radu2016towards} designed a multi-modal RBM where an RBM is constructed for each sensor modality, then the output of all the modalities are unified. \citep{li2016deep} added pooling after the fully-connected layers to extract the important features. \citep{fang2014recognizing} used a contrastive gradient (CG) method to update the weight in fine-tuning, which helps the network to search and convergence quickly in all directions. \citep{zhang2015real} further implemented RBM on a mobile phone for offline training, indicating RBM can be very light-weight. Similar to autoencoder, RBM/DBN can also perform unsupervised feature learning for HAR.

\subsection{Recurrent Neural Network}
Recurrent neural network~(RNN) is widely used in speech recognition and natural language processing by utilizing the temporal correlations between neurons. LSTM (long-short term memory) cells are often combined with RNN where LSTM is serving as the \textit{memory} units through gradient descent.

Few work used RNN for the HAR tasks \citep{hammerla2016deep,inoue2016deep,edel2016binarized,guan2017ensembles}, where the learning speed and resource consumption are the main concerns for HAR. \citep{inoue2016deep} investigated several model parameters first and then proposed a relatively \textit{good} model which can perform HAR with high throughput. \citep{edel2016binarized} proposed a binarized-BLSTM-RNN model, in which the weight parameters, input, and output of all hidden layers are all binary values. The main line of RNN based HAR models is dealing with resource-constrained environments while still achieve good performance.

\begin{table*}[htbp]
	\centering
	\caption{Public HAR datasets (A=accelerometer, G=gyroscope, M=magnetometer, O=object sensor, AM=ambient sensor, ECG=electrocardiograph)}
	\label{tb-dataset}
	\resizebox{1\textwidth}{!}{
		\begin{tabular}{|c|c|c|c|c|c|c|c|c|}
			\hline
			\textbf{ID} & \textbf{Dataset} & \textbf{Type} & \textbf{\#Subject} & \textbf{S. Rate} & \textbf{\#Activity} & \textbf{\#Sample} & \textbf{Sensor} & \textbf{Reference} \\ \hline \hline
			D01 & OPPORTUNITY & ADL & 4 & 32 Hz & 16 & 701,366 & A, G, M, O, AM & \citep{ordonez2016deep} \\ \hline
			D02 & Skoda Checkpoint & Factory & 1 & 96 Hz & 10 & 22,000 & A & \citep{plotz2011feature} \\ \hline
			D03 & UCI Smartphone & ADL & 30 & 50 Hz & 6 & 10,299 & A, G & \citep{almaslukh2017effective} \\ \hline
			D04 & PAMAP2 & ADL & 9 & 100 Hz & 18 & 2,844,868 & A, G, M & \citep{zheng2014time} \\ \hline
			D05 & USC-HAD & ADL & 14 & 100 Hz & 12 & 2,520,000 & A, G & \citep{jiang2015human} \\ \hline
			D06 & WISDM & ADL & 29 & 20 Hz & 6 & 1,098,207 & A & \citep{alsheikh2015deep} \\ \hline
			D07 & DSADS & ADL & 8 & 25 Hz & 19 & 1,140,000 & A, G, M & \citep{zhang2015recognizing} \\ \hline
			D08 & Ambient kitchen & \small{Food preparation} & 20 & 40 Hz & 2 & 55,000 & O & \citep{plotz2011feature} \\ \hline
			D09 & \small{Darmstadt Daily Routines} & ADL & 1 & 100 Hz & 35 & 24,000 & A & \citep{plotz2011feature} \\ \hline
			D10 & Actitracker & ADL & 36 & 20 Hz & 6 & 2,980,765 & A & \citep{zeng2014convolutional} \\ \hline
			D11 & SHO & ADL & 10 & 50 Hz & 7 & 630,000 & A, G, M & \citep{jiang2015human} \\ \hline
			D12 & BIDMC & \small{Heart failure} & 15 & 125 Hz & 2 & \textgreater 20,000 & ECG & \citep{zheng2014time} \\ \hline
			D13 & MHEALTH & ADL & 10 & 50 Hz & 12 & 16,740 & A, C, G & \citep{ha2016convolutional} \\ \hline
			D14 & Daphnet Gait & Gait & 10 & 64 Hz & 2 & 1,917,887 & A & \citep{hammerla2016deep} \\ \hline
			D15 & ActiveMiles & ADL & 10 & 50-200 Hz & 7 & 4,390,726 & A & \citep{ravi2017deep} \\ \hline
			D16 & HASC & ADL & 1 & 200 Hz & 13 & NA & A & \citep{hayashi2015daily} \\ \hline
			D17 & PAF & PAF & 48 & 128 Hz & 2 & 230,400 & EEG & \citep{pourbabaee2017deep} \\ \hline
			D18 & ActRecTut & Gesture & 2 & 32 Hz & 12 & 102,613 & A, G & \citep{yang2015deep} \\ \hline
			D19 & Heterogeneous & ADL & 9 & 100-200 Hz & 6 & 43,930,257 & A, G & \citep{yao2017deepsense} \\ \hline
		\end{tabular}
	}
	\vspace{-.1in}
\end{table*}

\subsection{Hybrid Model}

Hybrid model is the combination of some deep models. 

%Since CNN is competent for feature extraction, some work combined CNN with other types of models. 

One emerging hybrid model is the combination of CNN and RNN. \citep{ordonez2016deep,yao2017deepsense} provided good examples for how to combine CNN and RNN. It is shown in~\citep{ordonez2016deep} that the performance of `CNN + recurrent dense layers' is better than `CNN + dense layers'. Similar results are also shown in~\citep{singh2017transforming}. The reason is that CNN is able to capture the spatial relationship, while RNN can make use of the temporal relationship. Combining CNN and RNN could enhance the ability to recognize different activities that have varied time span and signal distributions. Other work combined CNN with models such as SAE~\citep{zheng2016exploiting} and RBM~\citep{liu2016lasagna}. In those work, CNN performs feature extraction, and the generative models can help in speeding up the training process. In the future, we expect there will be more research in this area.

\section{Applications}
\label{sec-app}

HAR is always not the final goal of an application, but it serves as an important step in many applications such as skill assessment and smart home assistant. In this section, we survey deep learning based HAR from the application perspective. 

\subsection{Featured Applications}

Most of the surveyed work focused on recognizing \textit{activities of daily living} (ADL) and \textit{sports}~\citep{zeng2014convolutional,chen2015deep,ronao2016human,ravi2017deep}. Those activities of simple movements are easily captured by body-worn sensors. Some research studied people's \textit{lifestyle} such as sleep~\citep{sathyanarayana2016impact} and respiration~\citep{khan2017deep,hannink2017sensor}. The detection of such activities often requires some object and ambient sensors such as WiFi and sound, which are rather different from ADL. 

It is a developing trend to apply HAR to \textit{health and disease} issues. Some pioneering work has been done with regard to Parkinson's disease~\citep{hammerla2015pd}, trauma resuscitation~\citep{li2016deep,li2016deep2} and paroxysmal atrial fibrillation~(PAF)~\citep{pourbabaee2017deep}. Disease issues are always related to the change of certain body movements or functions, so they can be detected using corresponding sensors. 

Under those circumstances, the association between disease and activity should be given more consideration. It is important to use the appropriate sensors. For instance, Parkinson's disease is often related to the frozen of gait, which can be reflected by some inertial sensors attached to shoes~\citep{hammerla2015pd}.

Other than health and disease, the recognition of \textit{high-level} activities is helpful to learn more resourceful information for HAR. The movement, behavior, environment, emotion, and thought are critical parts in recognizing high-level activities. However, most work only focused on body movements in smart homes~\citep{vepakomma2015wristocracy,fang2014recognizing}, which is not enough to recognize high-level activities. For instance, \citep{vepakomma2015wristocracy} combined activity and environment signal to recognize activities in a smart home, but the activities are constrained to body movements without more information on user emotion and state, which are also important. In the future, we expect there will be more research in this area.

\subsection{Benchmark Datasets}
We extensively explore the benchmark datasets for deep learning based HAR. Basically, there are two types of data acquisition schemes: \textit{self data collection} and \textit{public datasets}.

\begin{itemize}[noitemsep,nolistsep]
	\item \textit{Self data collection:} Some work performed their own data collection~(e.g.~\citep{chen2015deep,zhang2015real,bhattacharya2016smart,zhang2015human}). Very detailed efforts are required for self data collection, and it is rather tedious to process the collected data.
	
	\item \textit{Public datasets:} There are already many public HAR datasets that are adopted by most researchers~(e.g.~\citep{plotz2011feature,ravi2016deep,hammerla2016deep}). By summarizing existing literature, we present several widely used public datasets in Table~\ref{tb-dataset}. 
\end{itemize}

%Adopting public datasets can save the time for data collection and preprocessing, and protocols from previous work can be easily used as comparison.

\begin{table*}[!p]
	\centering
	\caption{Summation of existing works based on the three aspects: sensor modality, deep model and application (in literature order)}
	\vspace{.1in}
	\label{tb-all}
	\resizebox{1\textwidth}{!}{
		\begin{tabular}{|c|c|c|c|c|}
			\hline
			\textbf{Literature} & \textbf{Sensor Modality} & \textbf{Deep Model} & \textbf{Application} & \textbf{Dataset}\\ \hline \hline
			\citep{almaslukh2017effective} & Body-worn & SAE & ADL & D03\\ \hline
			\citep{alsheikh2015deep} & Body-worn & RBM & ADL, factory, Parkinson & D02, D06, D14\\ \hline
			\citep{bhattacharya2016smart} & Body-worn, ambiemt & RBM & Gesture, ADL, transportation & Self, D01\\ \hline
			\citep{chen2015deep} & Body-worn & CNN & ADL & Self  \\ \hline
			\citep{chen2016lstm} & Body-worn & CNN & ADL & D06  \\ \hline
			\citep{cheng2017human} & Body-worn & DNN & Parkinson & Self \\ \hline
			\citep{edel2016binarized} & Body-worn & RNN & ADL & D01, D04, Self\\ \hline
			\citep{fang2014recognizing} & Object, ambient & DBN & ADL & Self \\ \hline
			\citep{gjoreskicomparing} & Body-worn & CNN & ADL & Self, D01\\ \hline
			\citep{guan2017ensembles} & Body-worn, object, ambient & RNN & ADL, smart home & D01, D02, D04 \\ \hline
			\citep{ha2015multi} & Body-worn & CNN & Factory, health & D02, D13\\ \hline
			\citep{ha2016convolutional} & Body-worn & CNN & ADL, health & D13\\ \hline
			\citep{hammerla2015pd} & Body-worn & RBM & Parkinson & Self\\ \hline
			\citep{hammerla2016deep} & Body-worn, object, ambient & DNN, CNN, RNN & ADL, smart home, gait & D01, D04, D14 \\ \hline
			\citep{hannink2017sensor} & Body-worn & CNN & Gait & Self\\ \hline
			\citep{hayashi2015daily} & Body-worn, ambient & RBM & ADL, smart home & D16 \\ \hline
			\citep{inoue2016deep} & Body-worn & RNN & ADL & D16 \\ \hline
			\citep{jiang2015human} & Body-worn & CNN & ADL & D03, D05, D11\\ \hline
			\citep{khan2017deep} & Ambient & CNN & Respiration & Self \\ \hline
			\citep{kim2016hand} & Ambient & CNN & Hand gesture & Self \\ \hline
			\citep{kim2017human} & Body-worn & CNN & ADL & Self\\ \hline
			\citep{lane2015can} & Body-worn, ambient & RBM & ADL, emotion  & Self\\ \hline
			\citep{lane2015deepear} & Ambient & RBM & ADL & Self \\ \hline
			\citep{lee2017human} & Body-worn & CNN & ADL & Self\\ \hline
			\citep{li2016deep} & Object & RBM & Patient resuscitation & Self \\ \hline
			\citep{li2016deep2} & Object & CNN & Patient resuscitation & Self \\ \hline
			\citep{li2014unsupervised} & Body-worn & SAE & ADL & D03 \\ \hline
			\citep{liu2016lasagna} & Body-worn & CNN, RBM & ADL  & Self \\ \hline
			\citep{mohammed2017unsupervised} & Body-worn & CNN & ADL, gesture & Self \\ \hline
			\citep{morales2016deep} & Body-worn & CNN & ADL, smart home & D01, D02 \\ \hline
			\citep{murad2017deep} & Body-worn & RNN & ADL, smart home & D01, D02, D05, D14 \\ \hline
			\citep{ordonez2016deep} & Body-worn & CNN, RNN & ADL, gesture, posture, factory & D01, D02 \\ \hline
			\citep{panwar2017cnn} & Body-worn & CNN & ADL & Self \\ \hline
			\citep{plotz2011feature} & Body-worn, object & RBM & ADL, food preparation, factory  & D01, D02, D08, D14 \\ \hline
			\citep{pourbabaee2017deep} & Body-worn & CNN & PAF disease & D17 \\ \hline
			\citep{radu2016towards} & Body-worn & RBM & ADL & D19\\ \hline
			\citep{ravi2016deep} & Body-worn & CNN & ADL, factory & D02, D06, D14, D15\\ \hline
			\citep{ravi2017deep} & Body-worn & CNN & ADL, factory, Parkinson & D02, D06, D14, D15\\ \hline
			\citep{ronao2015deep,ronao2015evaluation,ronao2016human} & Body-worn & CNN & ADL & D03\\ \hline
			\citep{sathyanarayana2016impact} & Body-worn & CNN, RNN, DNN & ADL, sleep & Self \\ \hline
			\citep{singh2017transforming} & Ambient & CNN, RNN & Gait & NA \\ \hline
			\citep{vepakomma2015wristocracy} & Body-worn, object, ambient & DNN & ADL  & Self\\ \hline
			\citep{walse2016pca} & Body-worn & DNN & ADL & D03\\ \hline
			\citep{wang2016device} & Body-worn, ambient & CNN & ADL, location & Self \\ \hline
			\citep{wang2016human} & Object, ambient & SAE & ADL & NA \\ \hline
			\citep{yang2015deep} & Body-worn, object, ambient & CNN & ADL, smart home, gesture & D01, D18 \\ \hline
			\citep{yao2017deepsense} & Body-worn, object & CNN, RNN & Cartrack, ADL & Self, D19 \\ \hline
			\citep{zebin2016human} & Body-worn & CNN & ADL & Self\\ \hline
			\citep{zeng2014convolutional} & Body-worn, ambient, object & CNN & ADL, smart home, factory & D01, D02, D10\\ \hline
			\citep{zhang2015human} & Body-worn & DNN & ADL & Self\\ \hline
			\citep{zhang2015real} & Body-worn & RBM & ADL & Self\\ \hline
			\citep{zhang2015recognizing} & Body-worn & DBN & ADL, smart home & D01, D05, D07\\ \hline
			\citep{zhang2017car} & Object & CNN & Medical & Self \\ \hline
			\citep{zhang2017human} & Body-worn & DNN & ADL & Self \\ \hline
			\citep{zheng2016exploiting} & Body-worn & CNN, SAE & ADL & D04\\ \hline
			\citep{zheng2014time} & Body-worn & CNN & ADL, heart failure & D04, D14\\ \hline
		\end{tabular}
	}
\end{table*}

\section{Summary and Discussion}
\label{sec-summary}

Table~\ref{tb-all} presents all the surveyed work in this article. We can make several observations based on the table.

\textbf{1) Sensor deployment and preprocessing.} Choosing the suitable sensors is critical for successful HAR. In surveyed literature, body-worn sensors serve as the most common modalities and accelerometer is mostly used. The reasons are two folds. Firstly, a lot of wearable devices such as smartphones or watches are equipped with an accelerometer, which is easy to access. Secondly, the accelerometer is competent to recognize many types of daily activities since most of them are simple body movements. Compared to body-worn sensors, object and ambient sensors are better at recognizing activities related to context and environment such as \textit{having coffee}. Therefore, it is suggested to use body-worn sensors~(mostly accelerometer+gyroscope) for ADL and sports activities. If the activities are pertaining to some semantic meaning but more than simple body movements, it is better to combine the object and ambient sensors. In addition, there are few public datasets for object and ambient sensors probably because of privacy issues and deployment difficulty of the data collecting system. We expect there will be more open datasets regarding those sensors.

Sensor placement is also important. Most body-worn sensors are placed on the dominant wrist, waist, and the dominant hip pocket. This placement strategy can help to recognize most common daily activities. However, when it comes to object and ambient sensors, it is critical to deploy them in a non-invasive way. Those sensors are not usually interacting with users directly, so it is critical to collect the data naturally and non-invasively.

Before using deep models, the raw sensor data need to be preprocessed accordingly. There are two important aspects. The first aspect is \textit{sliding window}. The inputs should be cut into individual inputs according to the sampling rate. This procedure is similar to conventional PR approaches. The second one is \textit{channels}. Different sensor modalities can be treated as separate channels, and each axis of a sensor can also be a channel. Using multi-channel could enhance the representation capability of the deep model since it can reflect the hidden knowledge of the sensor inputs.

\textbf{2) Model selection.} There are several deep models surveyed in this article. Then, a natural question arises: \textit{which model is the best for HAR?} \citep{hammerla2016deep} did an early work by investigating the performance of DNN, CNN and RNN through 4,000 experiments on some public HAR datasets. We combine their work and our explorations to draw some conclusions: RNN and LSTM are recommended to recognize short activities that have natural order while CNN is better at inferring long-term repetitive activities~\citep{hammerla2016deep}. The reason is that RNN could make use of the time-order relationship between sensor readings, and CNN is more capable of learning deep features contained in recursive patterns. For multi-modal signals, it is better to use CNN since the features can be integrated through multi-channel convolutions~\citep{zeng2014convolutional,zheng2014time,ha2015multi}. While adapting CNN, data-driven approaches are better than model-driven approaches as the inner properties of the activity signal can be exploited better when the input data are transformed into the virtual image. Multiple convolutions and poolings also help CNN perform better. RBM and autoencoders are usually pre-trained before being fine-tuned. Multi-layer RBM or SAE is preferred for more accurate recognition. 

Technically there is no model which outperforms all the others in all situations, so it is recommended to choose models based on the scenarios. To better illustrate the performance of some deep models, Table~\ref{tb-compare} offers some results comparison of existing work on public datasets in Table~\ref{tb-dataset}~\footnote{OPP 1, OPP 2, Skoda, and UCI smartphone follow the protocols in~\citep{hammerla2016deep}, \citep{plotz2011feature}, \citep{zeng2014convolutional}, and \citep{ronao2016human}, respectively. OPP 1 used weighted f1-score; OPP~2, Skoda, and UCI smartphone used accuracy.}. In Skoda and UCI Smartphone protocols, CNN achieves the best performance. In two OPPORTUNITY protocols, DBN and RNN outperform the others. This confirms that no models can achieve the best in all tasks. Moreover, the hybrid models tend to perform better than single models (DeepConvLSTM in OPPORTUNITY 1 and Skoda). For a single model, CNN with shifted inputs (Fourier transform) generates better results compared to shifted kernels.

%\textbf{3) Learning and optimization.} Learning is important in deep models. Some work studied learning rate. \citep{fang2014recognizing} used a contrastive gradient (CG) method to update the weight in fine-tuning, which helps the network to search and convergence quickly in all directions. There are several factors pertaining to learning and optimization. 1) \textit{learning strategy}. Stochastic gradient decent (SGD) is usually required to train a neural network, where some accelerating strategies such as momentum could improve the efficacy of training. 2) \textit{regularization}. \citep{hammerla2016deep} proposed a new regularization framework \textit{fANOVA}. Dropout is often required to avoid overfitting. When applying dropout, it is recommended to retain nodes with a lower probability (0.1 $\sim$ 0.3) for the shallow layers, while a higher probability (0.4 $\sim$ 0.5) for the higher layers (probably dense layers). Weight-decay should also be considered. 3) the \textit{learning rate}. A relatively smaller learning rate tends to perform better~\citep{ronao2015deep}. 4) \textit{activation function}. Most work adopted the ReLU (rectified linear unit) as the activation function and softmax is adopted for the last layer, while in RNN it is better to use hyperbolic functions. 5) \textit{hyperparameter search}. It is recommended to perform grid search to find the best hyperparameters. \textcolor{red}{(you man want to remove this paragraph)}

\begin{table}[!t]
	\centering
	\caption{Performance comparison of existing deep models}
	
	\label{tb-compare}
	\resizebox{0.48\textwidth}{!}{
		\begin{tabular}{llll}
			\hline
			\textbf{Protocol} & \textbf{Model} & \textbf{Result} & \textbf{Reference} \\ \hline
			\multirow{4}{1cm}{\textbf{OPP 1}} & {\small b-LSTM-S} & \textbf{92.70} & \citep{hammerla2016deep} \\ \cmidrule(l){2-4} 
			& CNN & 85.10 & \citep{yang2015deep} \\ \cmidrule(l){2-4} 
			& CNN & 88.30 & \citep{ordonez2016deep} \\ \cmidrule(l){2-4} 
			& {\scriptsize DeepConvLSTM} & 91.70 & \citep{ordonez2016deep} \\ \hline
			\multirow{3}{*}{\textbf{OPP 2}} & DBN & 73.20 & \citep{plotz2011feature} \\ \cmidrule(l){2-4} 
			& CNN & 76.80 & \citep{zeng2014convolutional} \\ \cmidrule(l){2-4} 
			& DBN & \textbf{83.30} & \citep{zhang2015recognizing} \\ \hline
			\multirow{3}{*}{\textbf{Skoda}} & CNN & 86.10 & \citep{zeng2014convolutional} \\ \cmidrule(l){2-4} 
			& CNN & 89.30 & \citep{alsheikh2015deep} \\ \cmidrule(l){2-4} 
			& {\scriptsize DeepConvLSTM} & \textbf{95.80} & \citep{ordonez2016deep} \\ \hline
			\multirow{4}{*}{\textbf{UCI smartphone}} & CNN & 94.61 & \citep{ronao2016human}  \\ \cmidrule(l){2-4} 
			& CNN & \textbf{95.18} & \citep{jiang2015human} \\ \cmidrule(l){2-4} 
			& CNN & 94.79 & \citep{ronao2015deep} \\ \cmidrule(l){2-4} 
			& CNN & 90.00 & \citep{ronao2015evaluation}  \\ \hline
		\end{tabular}
	}
	\vspace{-.2in}
\end{table}
%\footnotetext{OPP 1, OPP 2, Skoda, and UCI smartphone follow the protocols in~\citep{hammerla2016deep}, \citep{plotz2011feature}, \citep{zeng2014convolutional}, and \citep{ronao2016human}, respectively. OPP 1 used weighted f1-score; OPP~2, Skoda, and UCI smartphone used accuracy.}

\section{Grand Challenges}
\label{sec-discuss}
Despite the progress in previous work, there are still challenges for deep learning based HAR. In this section, we present those challenges and propose some feasible solutions.

\textbf{A. Online and mobile deep activity recognition.} 
Two critical issues are related to deep HAR: online deployment and mobile application. Although some existing work adopted deep HAR on smartphone~\citep{lane2015deepear} and watch~\citep{bhattacharya2016smart}, they are still far from online and mobile deployment. Because the model is often trained offline on some remote server and the mobile device only utilizes a trained model. This approach is neither real-time nor friendly to incremental learning. There are two approaches to tackle this problem: \textit{reducing the communication cost between mobile and server}, and \textit{enhancing computing ability of the mobile devices}.

\textbf{B. More accurate unsupervised activity recognition.} 
The performance of deep learning still relies heavily on labeled samples. Acquiring sufficient activity labels is expensive and time-consuming. Thus, \textit{unsupervised} activity recognition is urgent.
\begin{itemize}[noitemsep,nolistsep]
	\item \textit{Take advantage of the crowd.} The latest research indicates that exploiting the knowledge from the crowd will facilitate the task \citep{prelec2017solution}. Crowd-sourcing takes advantage of the crowd to annotate the unlabeled activities. Other than acquiring labels passively, researchers could also develop more elaborate, privacy-concerned way to collect useful labels.
	
	\item \textit{Deep transfer learning.} Transfer learning performs data annotation by leveraging labeled data from other auxiliary domains \citep{pan2010survey,cook2013transfer,wang2017balanced}. There are many factors related to human activity, which can be exploited as auxiliary information using deep transfer learning. Problems such as sharing weights between networks, exploiting knowledge between activity related domains, and how to find more relevant domains are to be resolved.
\end{itemize}

\textbf{C. Flexible models to recognize high-level activities.} 
More complex high-level activities need to be recognized other than only simple daily activities. It is difficult to determine the hierarchical structure of high-level activities because they contain more semantic and context information. Existing methods often ignore the correlation between signals, thus they cannot obtain good results.

\begin{itemize}[noitemsep,nolistsep]
	\item \textit{Hybrid sensor}.
	Elaborate information provided by the hybrid sensor is useful for recognizing fine-grained activities \citep{vepakomma2015wristocracy}. Special attention should be paid to the recognition of fine-grained activities by exploiting the collaboration of hybrid sensors.
	
	\item \textit{Exploit context information}.
	Context is any information that can be used to characterize the situation of an entity \citep{abowd1999towards}. Context information such as Wi-Fi, Bluetooth, and GPS can be used to infer more environmental knowledge about the activity. The exploitation of resourceful context information will greatly help to recognize user state as well as more specific activities. 
\end{itemize}

\textbf{D. Light-weight deep models.}
Deep models often require lots of computing resources, which is not available for wearable devices. In addition, the models are often trained off-line which cannot be executed in real-time. However, less complex models such as shallow NN and conventional PR methods could not achieve good performance. Therefore, it is necessary to develop light-weight deep models to perform HAR.

\begin{itemize}[noitemsep,nolistsep]
	\item \textit{Combination of human-crafted and deep features}. 
	Recent work indicated that human-crafted and deep features together could achieve better performance~\citep{plotz2011feature}. Some pre-knowledge about the activity will greatly contribute to more robust feature learning in deep models~\citep{stewart2017label}. Researchers should consider the possibility of applying two kinds of features to HAR with human experience and machine intelligence.
	
	\item \textit{Collaboration of deep and shallow models.} 
	Deep models have powerful learning abilities, while shallow models are more efficient. The collaboration of those two models has the potential to perform both accurate and light-weight HAR. Several issues such as how to share the parameters between deep and shallow models are to be addressed.
\end{itemize}

\textbf{E. Non-invasive activity sensing.} 
Traditional activity collection strategies need to be updated with more non-invasive approaches. Non-invasive approaches tend to collect information and infer activity without disturbing the subjects and requires more flexible computing resources.

\begin{itemize}[noitemsep,nolistsep]
	\item \textit{Opportunistic activity sensing with deep learning.} Opportunistic sensing could dynamically harness the non-continuous activity signal to accomplish activity inference~\citep{chen2016ocean}. In this scenario, back propagation of deep models should be well-designed. 
\end{itemize}

\textbf{F. Beyond activity recognition: assessment and assistant.} 
Recognizing activities is often the initial step in many applications. For instance, some professional skill assessment is required in fitness exercises and smart home assistant plays an important role in healthcare services. There is some early work on climbing assessment~\citep{khan2015beyond}. With the advancement of deep learning, more applications should be developed to be beyond just recognition.

%\subsection{Context}
%Another critical problem we can notice from the surveyed work is, there are almost no particular research that utilized context signals to perform wAR. In fact it is rather difficult to define context, we simply take the 
%definition from : 
%
%\subsection{Online Learning}
%Deep learning requires highly resourceful computing environment, yet there are only limited battery computing capability on the wearable devices such as smartphones and smart watches. For literatures that performed data collection on wearable devices, most of them were performing the deep model building on the server end instead of on the device end. However, it seems that a highly scalable, wearable deployment, where effective deep models can directly run on the device is the future.
%
%Few researchers made trials. In \citep{}, authors implemented an SoC-level RBM model on a smart watch. \citep{zhang2015real} collected activity data on a Samsung smartphone, then after online model-training, authors fed the parameters from DBN model to the device. However, this work is highly dependent on the off-line training results and is not generalizable. In contrast, \citep{lane2015deepear} implemented a DNN model on an Android 4.3 smartphone by partially modifying the DSP module. To our knowledge, it is the first work to implement deep audio sensing on a mobile platform. Besides, \citep{lane2016deepx} built a software accelerator called \textit{DeepX} to assist the mobile deployment of deep learning moducles.

\section{Conclusion}
\label{sec-conclusion}
Human activity recognition is an important research topic in pattern recognition and pervasive computing. In this paper, we survey the recent advance in deep learning approaches for sensor-based activity recognition. Compared to traditional pattern recognition methods, deep learning reduces the dependency on human-crafted feature extraction and achieves better performance by automatically learning high-level representations of the sensor data. We highlight the recent progress in three important categories: sensor modality, deep model, and application. Subsequently, we summarize and discuss the surveyed research in detail. Finally, several grand challenges and feasible solutions are presented for future research.

\section*{Acknowledgments}
This work is supported in part by National Key R \& D Program of China~(No.2017YFB1002801), NSFC~(No.61572471), and Science and Technology Planning Project of Guangdong Province~(No.2015B010105001). Authors thank the reviewers for their valuable comments.

\bibliographystyle{model2-names}
\bibliography{refs}

\end{document}